\documentclass[graybox]{svmult}

\usepackage{mathptmx}       
\usepackage{helvet}         
\usepackage{courier}        
\usepackage{type1cm}        
\usepackage{makeidx}         
\usepackage{graphicx}        
\graphicspath{{Figures/}}
\usepackage{multicol}        
\usepackage[bottom]{footmisc}

\usepackage{hyperref}

\makeindex             

\begin{document}

\title*{Robot Accident Investigation: a case study in Responsible Robotics}
\titlerunning{Robot Accident Investigation}
\author{Alan F.T. Winfield, Katie Winkle, Helena Webb, Ulrik Lyngs, Marina Jirotka and Carl Macrae}
\authorrunning{Winfield, Winkle, Webb, Lyngs, Jirotka and Macrae} 
\institute{A Winfield, K Winkle \at Bristol Robotics Lab, UWE Bristol, UK. \email{alan.winfield@brl.ac.uk, katie.winkle@brl.ac.uk }
\and H Webb, U Lyngs, M Jirotka \at Department of Computer Science, University of Oxford, UK. \email{helena.webb@cs.ox.ac.uk, ulrik.lyngs@cs.ox.ac.uk, marina.jirotka@cs.ox.ac.uk}
\and C Macrae \at Nottingham University Business School, University of Nottingham, UK. \email{Carl.Macrae@nottingham.ac.uk}
}
%
%
\maketitle


\abstract{Robot accidents are inevitable. Although rare, they have been happening since assembly-line robots were first introduced in the 1960s. But a new generation of social robots are now becoming commonplace. Often with sophisticated embedded artificial intelligence (AI) social robots might be deployed as care robots to assist elderly or disabled people to live independently. Smart robot toys offer a compelling interactive play experience for children and increasingly capable autonomous vehicles (AVs) the promise of hands-free personal transport and fully autonomous taxis. Unlike industrial robots which are deployed in safety cages, social robots are designed to operate in human environments and interact closely with humans; the likelihood of robot accidents is therefore much greater for social robots than industrial robots. This paper sets out a draft framework for social robot accident investigation; a framework which proposes both the technology and processes that would allow social robot accidents to be investigated with no less rigour than we expect of air or rail accident investigations. The paper also places accident investigation within the practice of responsible robotics, and makes the case that social robotics without accident investigation would be no less irresponsible than aviation without air accident investigation.}

\section{Introduction}
\label{sec:1}
\begin{svgraybox}
\textit{What could possibly go wrong?}
\newline
Imagine that your elderly mother, or grandmother, has an assisted living robot to help her live independently at home. The robot is capable of fetching her drinks, reminding her to take her medicine and keeping in touch with family. Then one afternoon you get a call from a neighbour who has called round and sees your grandmother collapsed on the floor. When the paramedics arrive they find the robot wandering around apparently aimlessly. One of its functions is to call for help if your grandmother falls or stops moving, but it seems that the robot failed to do this.
\setlength{\leftskip}{8mm}

Fortunately your grandmother makes a full recovery. Not surprisingly you want to know what happened: did the robot cause the accident? Or maybe it didn't but made matters worse, and why did it fail to raise the alarm?
\setlength{\leftskip}{8mm}
\end{svgraybox}
\setlength{\leftskip}{0pt}

Although this is a fictional scenario it could happen today. If it did we would be totally reliant on the goodwill of the robot manufacturer to discover what went wrong. Even then we might not get the answers we seek; it is entirely possible that neither the robot nor the company that made it are equipped with the tools and processes to undertake an investigation.

Robot accidents are inevitable. Although rare, they have been happening since assembly line robots were first introduced in the 1960s. First wave robots include factory robots (multi-axis manipulators), autonomous guided vehicles (AGVs) deployed in warehouses, Remotely Operated Vehicles (ROVs) for undersea exploration and maintenance, teleoperated robots for bomb disposal, and (perhaps surprisingly) Mars rovers for planetary exploration. A defining characteristic of first wave robots is that they are designed for jobs that are dull, dirty or dangerous; these robots are typically either dangerous to be around (and therefore enclosed in safety cages on assembly lines), or deployed in dangerous or inaccessible environments. 

In contrast second wave robots are designed to operate in human environments and interact directly with people. Those human environments include homes, offices, hospitals, shopping malls, and city or urban streets and -- unlike first wave robots -- many are designed to be used by untrained, naive or vulnerable users, including children and the elderly. These are robots in society, and hence social robots\footnote{Noting that we take a broader view of social robotics than usual.}. Often with sophisticated embedded artificial intelligence (AI) social robots might be deployed as care robots to assist elderly or disabled people to live independently. Smart robot toys offer a compelling interactive play experience for children and increasingly capable autonomous vehicles (AVs) the promise of hands-free personal transport and fully autonomous taxis.

Social robots by definition work with and alongside people in human environments, thus the likelihood and scope of robot accidents is much greater than with industrial robots. This is not just because of the close proximity of social robots and their users (and perhaps also bystanders), it is also because of the kinds of roles such robots are designed to fulfill, and further exacerbated by the unpredictability of people and the unstructured, dynamic nature of human environments.

Given the inevitability of social robot accidents it is perhaps surprising that no frameworks or processes of social robot accident investigation exist. This paper addresses that deficit by setting out a draft framework for social robot accident investigation; a framework which proposes both the technology and processes that would allow social robot accidents to be investigated with no less rigour than we expect of air or rail accident investigations.

This paper proceeds as follows. We first survey the current practices and frameworks for accident investigation, including in transport (air, rail and road) and in healthcare, in section 2.  In section 3 we survey robot accidents, including both industrial and social robot accidents, then analyse the scope for social robot accidents in order to understand why social robot accidents are more likely (per robot deployed) than industrial robot accidents. Section 4 then places accident investigation within the practice of responsible robotics, by defining responsible robotics within the broader practice of Responsible Innovation; the section also briefly surveys the emerging practices of values-driven design and ethical standards in robotics. Section 5 sets out both the technology and processes of our draft framework for social robot accident investigation, then illustrates the application of this framework by setting out how an investigation of our fictional accident might play out. Finally, in section 6, we conclude by outlining work currently underway within project RoboTIPS to develop and validate our framework with real-robot mock accident scenarios, before considering some key questions about who would investigate real-world social robot accidents.

\section{The practice of Accident Investigation}
\label{sec:2}
Investigating accidents and incidents is a routine and widespread activity in safety-critical industries such as aviation, the railways and healthcare. In healthcare, for instance, over 2 million safety incidents are reported across the English National Health Service each year, with around 50,000 of those incidents causing moderate to severe harm or death \cite{nhs19} -- such as medication errors or wrong-site surgery. Accident investigation also has a long history. In aviation, the world's first national air accident investigation agency was established over a century ago in 1915 \cite{aaib15}, and the international conventions governing air accident investigation were agreed in the middle of the last century \cite{icaa07}. The practices and methods of accident investigation are therefore well-established in many sectors and have been developed and refined over many years. As such, when considering a framework for the investigation of social robot accidents, it is instructive to examine how accident investigation is conducted in other safety-critical domains. 

First, it is important to emphasise that the core principle and fundamental purpose of accident investigation is learning: while investigations primarily focus on examining events that have occurred in the past, the core purpose of an accident investigation is to improve safety in the future. Accident investigations are therefore organised around three key questions \cite{mac17}. The first is factual: what actually happened? The second is explanatory: why did those things happen and what does that reveal about weaknesses in the design and operation of a particular system? The third is practical: how can systems be improved to prevent similar events happening in future? The ultimate objective for any accident investigation is to develop practical, achievable and targeted recommendations for making safety improvements. 

Conducting an accident investigation involves collecting and analysing a range of evidence and data from a variety of sources to understand what happened and why. This can include quantitative data, such as information from `black box' Flight Data Recorders on aircraft that automatically collect data on thousands of flight parameters. It can also include qualitative data in the form of organisational documentation and policies, and in-depth interviews with witnesses or those directly or indirectly involved in an accident such as a pilot or a maintenance engineer. Accident investigators therefore need to be experts in the methods and processes of accident investigation, and also need to have a deep and broad understanding of the industry and the technologies involved. However, accident investigations are also typically collaborative processes that are conducted by a diverse team of investigators who, in turn, draw on specific sources of expertise where required \cite{mac10}. 

A variety of methods have been developed to assist in the collection and analysis of safety data, from cognitive interviewing techniques \cite{fish92} to detailed human factors methods \cite{stant13} and organisational and sociotechnical systems analysis techniques \cite{under14}. Importantly, to understand why an accident has occurred and to determine how safety can be improved in future, accident investigations typically apply a systemic model of safety that helps identify, organise and explain the factors involved and how they are interconnected. One of the most widely applied and practical accident models is the ‘organisational accident’ framework -- commonly referred to as the ‘Swiss Cheese’ model \cite{reas97}. This has been adapted and applied in various ways \cite{atsb07}, but at core this provides a simple framework that conceptualises system safety as dependent on multiple layers of risk controls and safety defences that span from the front-line to organisational and regulatory levels -- such as redundant systems, emergency alarms, operator training, management decisions or regulatory requirements. Each of these safety defences will inevitably be partial or weak in some way, and accidents occur when the ‘holes’ in these defences all line up in some unexpected way -- thus the eponymous image of multiple slices of ‘Swiss Cheese’, each slice full of holes. The focus of accident investigations is to examine the entire sociotechnical system to understand the safety defences, the weaknesses in those defences, why those weaknesses arose and how they can be addressed. Similar premises underpin a variety of different accident investigation models, methods and tools. 
 
Safety-critical industries have developed sophisticated systems to support these activities of investigation and learning at all levels. These range from lengthy investigations of major accidents that are coordinated by national investigative bodies to relatively rapid local-level investigations of more minor incidents or near-miss events that are conducted by investigators within individual organisations \cite{mac14}. A lot of media attention understandably focuses on the investigations into high-profile accidents that are conducted by national investigation bodies, such as the US National Transportation Safety Board's investigations of the various accidents involving semi-automated Tesla vehicles \cite{ntsb16} and Uber's autonomous test vehicle \cite{ntsb18}. However, much of the more routine work of investigation actually occurs within individual organisations, like airlines and hospitals, which regularly conduct hundreds or thousands of investigations each year. These local-level investigations examine more minor safety incidents as well as near-miss events -- where there was no adverse outcome but some sort of safety-relevant failure was detected, such as a poorly specified maintenance procedure in an airline leading to a technical failure that causes a rejected take-off \cite{mac14}. Local-level investigations employ similar methods and approaches to those conducted at a national level, but are often much more rapid, lasting days or weeks rather than months and years. They are also typically limited to examining sources of risk within one single organisation, unlike national-level investigations which can consider regulatory factors and interactions between various different organisations across an industry. 

At all these levels of investigative activity, accident and incident investigation is always focused on learning. One of the main implications of this focus on learning is that accident investigation activities are typically entirely separated from other investigative activities that seek to attribute blame or determine liability. In aviation, for example, the information collected during major accident investigations may only be used for the purposes of safety investigation and improvement -- and may not be used, for instance, to punish individuals or pursue legal claims against organisations (EU 2010). This is also the case within individual organisations in safety-critical industries, where it is common to have a safety department or team that is entirely separated from operational processes or production pressures and whose sole focus is monitoring safety, investigating incidents and supporting improvement \cite{mac16}. The information that is collected by these safety teams is usually only used for safety improvement purposes and purposefully not used for line management or disciplinary processes, to ensure that staff feel they can openly and honestly provide safety-relevant information to support ongoing efforts to investigate and understand failures and continuously improve safety. 

\section{Robot Accidents}

Robert Williams is believed to be the first person killed by a robot in an industrial accident, in January 1979, at a Ford Motor Company casting plant\footnote{\url{https://en.wikipedia.org/wiki/Robert_Williams_(robot_fatality)}}. Given that there are over 2.4M industrial robots in use worldwide \cite{ifr19}, it is surprisingly difficult to find statistics on robot accidents. The US Department of Labor's Occupational Safety and Health Administration (OHSA) maintains a web page listing industrial accidents since 1984, and a search with the keyword `robot' returns records of 43 accidents\footnote{\url{https://www.osha.gov/pls/imis/AccidentSearch.search?acc_keyword=\%22Robot\%22&keyword_list=on}}; all resulted in injuries of which 29 were fatal. The US National Institute for Occupational Safety and Health (NIOSH) found 61 robot related deaths between 1992 and 2015, noting that ``These fatalities will likely increase over time because of the increasing number of conventional industrial robots being used by companies in the United States, and from the introduction of collaborative and co-existing robots, powered exoskeletons, and autonomous vehicles into the work environment.''\footnote{\url{https://www.cdc.gov/niosh/topics/robotics/aboutthecenter.html}} 

Finding data on accidents involving robots which interact with humans (HRI) is also difficult. One study on the safety of interactive industrial robots in Finland \cite{malm10} notes that ``International accident-report data related to robots is scarce''. The study reports that the majority of the 6000 robots in Finland are ``used in small robot work cells rather than large automation lines'' and that ``a natural feature of the production is that it requires substantial (mainly simple) HRI inside the robot working envelope''. The study analyses a total of 25 severe robot or manipulator-related accidents between 1989-2006, of which 3 were fatal, noting also that most of the accidents occurred toward the end of this period. Key characteristics of these accidents were: 

\begin{itemize}
\item ``The cause of an accident is usually a sequence of events, which have been difficult to foresee.
\item Operator inattentiveness and forgetfulness may lead them to misjudge a situation and even a normal robot movement may surprise them.
\item Most of the accidents involving robots occurred so that (the) robot moved unexpectedly (from worker's point of view) against the worker within the robot working area.
\item Inadequate safeguarding featured significantly as a cause of accidents.
\item Many accidents are associated with troubleshooting disturbances. And
\item only about 20\% of accidents occurred during undisturbed automated runs.'' \cite{malm10}
\end{itemize} 

Although now somewhat dated, Chapter 4 `Robot Accidents’ of \cite{dhil91} sets out a comprehensive analysis of industrial robot accidents, including data from accidents in Japan, Western Europe and the United States. Noting that ``human behavior plays an important role in certain robot accidents'' the paper outlines a set of typical human behaviours that might result in injury. A number of these are especially instructive to the present study:
\begin{itemize}
\item ``Humans often incorrectly read, or fail to see, instructions and labels on various products. 
\item Many people carry out most assigned tasks while thinking about other things. 
\item In most cases humans use their hands to test or examine. 
\item Many humans are unable to estimate speed, distance, or clearances very well. In fact, humans underestimate large distances and overestimate short distances. 
\item Many humans fail to take the time necessary to observe safety precautions. 
\item A sizeable portion of humans become complacent after a long successful exposure to dangerous items. 
\item There is only a small percentage of humans which recognize the fact that they cannot see well enough, either due to poor illumination or poor eyesight.'' \cite{dhil91}
\end{itemize}

Finding data on non-industrial robot accidents is even more difficult, but one study reports on adverse events in robotics surgery in the US \cite{alem16}; the paper surveys 10,624 reports collected by the Food and Drug Administration between 2000 and 2013, showing 8061 device malfunctions, 1391 patient injuries and 144 deaths, during a total of more than 1.75M procedures. Some robot accidents make headline news, for example car accidents in which semi-automated driver-assist functions are implicated; a total of six `self-driving car fatalities' have been reported since January 2016\footnote{\url{https://en.wikipedia.org/wiki/List_of_self-driving_car_fatalities}}. We only know of one accident in which a child's leg was bruised by a shopping mall security robot because it was reported in the national press\footnote{\url{https://www.wsj.com/articles/security-robot-suspended-after-colliding-with-a-toddler-1468446311}}.

A recent study in human-robot interaction examines several serious accidents, in aviation, the nuclear industry, and autonomous vehicles in an effort to understand ``the potential mismatches that can occur between control and responsibility in automated systems'' and argues that these mismatches ``create moral crumple zones, in which human operators take on the blame for errors or accidents not entirely in their control'' \cite{elish19}.

\subsection{The Scope for Social Robot Accidents}
\label{subsec:2}
As we have outlined above, industrial and surgical robot accidents are -- thankfully -- rare events. It is most unlikely that social robot accidents will be so rare. There are several factors that lead us to make this forecast:

\begin{enumerate}
\item Social robots, as broadly defined in this paper, are designed to be integrated into society at large: in urban and city environments, schools, hospitals, shopping malls, and in both care and private homes. Unlike industrial robots operating behind safety cages social robots are designed to be among us, up close and personal. It is this close proximity that undoubtedly presents the greatest risk.
\item Operators of industrial robots are required to undertake training courses, both on how to operate their robots and on the robot's safety features, and are expected to follow strict safety protocols. In contrast social robots are designed to benefit naive users -- including children, the elderly, and vulnerable or disabled people -- with little more preparation than might be expected to operate a vacuum cleaner or dishwasher.  
\item Industrial robots typically operate in an environment optimised for them and not humans. Social robots have no such advantage: humans (especially young or vulnerable humans) are unpredictable and human environments (from a robot’s perspective) are messy, unstructured and constantly changing. Designing robots capable of operating safely in human environments remains a serious challenge.
\item The range of serious harms that can arise from social robots is much greater than those associated with industrial robots. Social robots can, like their industrial counterparts, cause physical injury, but the responsible roboticist should be equally concerned about the potential for psychological harms, such as deception (i.e. a user coming to believe that a robot cares for them), over-trust or over-reliance on the robot, or violations of privacy or dignity. A number of \textit{ethical hazards}  are outlined in section 4 below\footnote{An ethical hazard is a possible source of ethical harm}. 
\item The scope of social robot accidents is thus enlarged. The nature of the roles social robots play will play in our lives -- supporting elderly people to live independently or helping the development of children with autism for instance \cite{broad17} -- opens up a range of ethical risks and vulnerabilities that have hitherto not been a concern of either robot designers or accident investigators. This factor also increases the likelihood of social robot accidents.
\end{enumerate}

If we are right and the near future brings an increasing number of social robot accidents, these accidents will need to be investigated in order to address the three key questions of accident investigation outlined in section 2. What happened? Why did it happen? And what must we change to ensure it doesn’t happen again? \cite{mac17}.

\section{Responsible Robotics}

In essence, Responsible Robotics is the application of Responsible Innovation (RI) to the field of robotics, so we first need to define RI. A straightforward definition of RI is: a set of good practices for ensuring that research and innovation benefits society and the environment. There are several frameworks for RI. One is the EPSRC AREA framework\footnote{\url{https://epsrc.ukri.org/research/framework/area/}}, built on the four pillars of Anticipation (of potential impacts and risks), Reflection (on the purposes of, motivations for and potential implications of the research), Engagement (to open up dialogue with stakeholders beyond the narrow research community), and Action (to use the three processes of anticipation, reflection and engagement, to influence the direction and trajectory of the research) \cite{owen14}. The more broadly framed Rome Declaration on Responsible Research and Innovation\footnote{\url{https://ec.europa.eu/research/swafs/pdf/rome_declaration_RRI_final_21_November.pdf}} is built around the six pillars of open access, governance, ethics, science communication, public engagement and gender equality\footnote{\url{http://ec.europa.eu/research/science-society/document_library/pdf_06/responsible-research-and-innovation-leaflet_en.pdf}}.

We define Responsible Robotics as follows:
\begin{svgraybox}
Responsible Robotics is the application of Responsible Innovation in the design, manufacture, operation, repair and end-of-life recycling of robots, that seeks the most benefit to society and the least harm to the environment.
\setlength{\leftskip}{8mm}
\end{svgraybox}
\setlength{\leftskip}{0pt}

Robot ethics -- which is concerned with the ethical impact of robots, on individuals, society and the environment, and how any negative impacts can be mitigated -- and ethical governance both have an important role in the practice of responsible robotics. In recent years many sets of ethical principles have been proposed in robotics and AI -- for a comprehensive survey see \cite{jobin19} -- but one of the earliest are the EPSRC Principles of Robotics, first published online in 2011\footnote{\url{https://epsrc.ukri.org/research/ourportfolio/themes/engineering/activities/principlesofrobotics/}}\cite{boden17}. In \cite{winf18} we set out a framework for ethical governance which links ethical principles to standards and regulation and argue that when such processes are robust and transparent, trust in robotics (or, to be precise, the individuals, companies and institutions designing, operating and regulating robots) will grow.  

Responsible social robotics \cite{webb19} is the practice of responsible robots in society with a particular ambition of creating robots which bring real benefits in both quality of life and safeguarding to the most vulnerable, within a framework of values-based design based upon a deep consideration of the ethical risks of social robots.  
There are a number of approaches and methods available to the responsible roboticist, including emerging new ethical standards, and an approach called ethically aligned design, which we will now review.

\subsection{Ethically Aligned Design}

In April 2016, the IEEE Standards Association launched a Global Initiative on the Ethics of Autonomous and Intelligent Systems\footnote{\url{https://standards.ieee.org/content/dam/ieee-standards/standards/web/documents/other/ec_about_us.pdf}}. The initiative positions human well-being as its central tenet. The initiative's mission is ``to ensure every stakeholder involved in the design and development of autonomous and intelligent systems (AIS) is educated, trained, and empowered to prioritize ethical considerations so that these technologies are advanced for the benefit of humanity''.

The first major output from the IEEE global ethics initiative is a document called Ethically Aligned Design (EAD) \cite{ieee19}. Developed through an iterative process over 3 years EAD is built on the three pillars of Universal Human Values, Political self-determination \& Data Agency, and Dependability, and eight general (ethical) principles covering Human Rights, Well-being, Data Agency, Effectiveness, Transparency, Accountability, Awareness of Misuse and Competence, and sets out more than 150 issues and recommendations across its 10 chapters. In essence EAD is both a manifesto and roadmap for a values-driven approach to the design of autonomous and intelligent systems. Spiekermann and Winkler \cite{spiek20} detail the process of ethically aligned design within a broader methodological framework they call value-based engineering for ethics by design. It is clear that responsible social robotics must be values-based.

\subsection{Standards in Social Robotics}

A foundational standard in social robotics is ISO 13482:2014 \textit{Safety requirements for personal care robots}. ISO 13482 covers mobile servant robots, physical assistant robots, and person carrier robots, but not robot toys or medical robots \cite{iso14}. The standard sets out a number of safety (but not ethical) hazards, including hazards relating to battery charging, robot motion, incorrect autonomous decisions and actions, and lack of awareness of robots by humans.

A new generation of explicitly ethical standards are now emerging \cite{winf19,odon20}. Standards are simply ``consensus-based agreed-upon ways of doing things'' \cite{bry17}. Although all standards embody a principle or value, explicitly ethical standards address clearly articulated ethical concerns and -- through their application -- seek to remove, reduce or highlight the potential for unethical impacts or their consequences \cite{winf19}.

The IEEE ethics initiative outlined above has initiated a total of 13 working groups to date, each tasked with drafting a new standard within the 7000 series of so-called `human standards'. The first of these to reach publication is IEEE 7010-2020 \textit{Recommended Practice for Assessing the Impact of Autonomous and Intelligent Systems on Human Well-Being}\footnote{\url{https://standards.ieee.org/standard/7010-2020.html}}.

Almost certainly the world’s first explicitly ethical standard in robotics is BS8611-2016 \textit{Guide to the ethical design and application of robots and robotic systems} \cite{bsi16}. ``BS8611 is not a code of practice, but instead guidance on how designers can undertake an ethical risk\footnote{An ethical risk is a possible ethical harm arising from exposure to a hazard.} assessment of their robot or system, and mitigate any ethical risks so identified. At its heart is a set of 20 distinct ethical hazards and risks, grouped under four categories: societal, application, commercial and financial, and environmental. Advice on measures to mitigate the impact of each risk is given, along with suggestions on how such measures might be verified or validated''\cite{winf19}. Societal hazards include, for example, anthropomorphisation, loss of trust, deception, infringements of privacy and confidentiality, addiction, and loss of employment, to which we should add the Uncanny Valley\cite{moor12}, weak security, lack of transparency (for instance the lack of data logs needed to investigate accidents), unrepairability and unrecyclability. Ethical risk assessment is a powerful and essential addition to the responsible roboticists toolkit, as it can be thought of as the opposite face of accident investigation, seeking -- at design time -- to prevent risks becoming accidents. 

\section{A Draft Framework for Social Robot Accident Investigation}

We now set out a framework for social robot accident investigation outlining first the technology, then the process, followed by an illustration of how the framework might be applied.

\subsection{Technology}

We have previously proposed that all robots should be equipped with the equivalent of an aircraft Flight Data Recorder (FDR) to continuously record sensor inputs, actuator outputs and relevant internal status data \cite{winf17}. We call this an ethical black box\footnote{Because it would be unethical not to have one.} (EBB), and argue that the EBB will play an important role in the process of discovering how and why a robot caused an accident.

\begin{figure}[ht]
\begin{center}

\includegraphics[scale=.5]{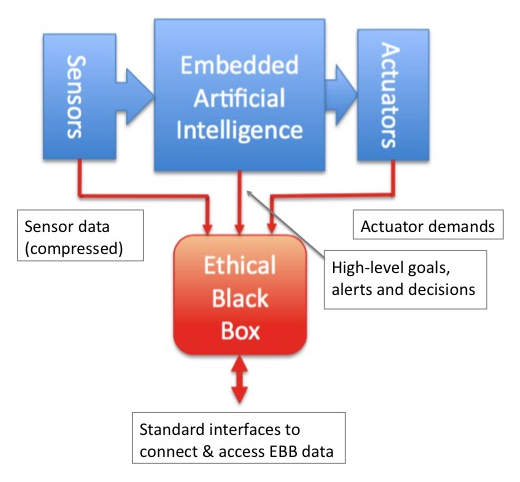}
\caption{Robot sub-systems with an Ethical Black Box and key dataflows} 
\label{fig:1}       
    
\end{center}
\end{figure}

All robots collect sense data, then -- on the basis of that sense data and some internal decision making process (the embedded Artificial Intelligence) -- send commands to actuators. This is of course a simplification of what in practice will be a complex set of connected systems and processes but, at an abstract level, all intelligent robots will have the three major subsystems shown in blue, in Fig. \ref{fig:1} Our EBB will have much in common with its aviation and automotive counterparts, the flight data recorder \cite{gros06} and event data recorder (EDR) \cite{gab04}, in particular: data is stored securely in a rugged unit designed to survive accidents; stored records are time and date stamped, and storage capacity is limited and organised such that only the most recent data are saved -- overwriting older records (an FDR typically records 17-25 hours of data, an automobile EDR as little as 30 seconds).

The technical specification for an EBB for a social robot is beyond the scope of this paper. It is, however, important to outline here the kinds of data we should expect to be recorded in the EBB. Consider the Pepper robot, as an exemplar of an advanced social robot \cite{pand18}. The Pepper is equipped with 25 sensors, including four microphones, two colour cameras, two accelerometers and various proximal and distal sensors. It has 20 motors, a chest mounted touch display pad, loudspeakers for speech output, and WiFi connectivity. We would therefore expect an EBB designed for the Pepper robot to be able to record:

\begin{enumerate}
\item Records of input sense data, including sampled (and compressed) camera images, audio sampled by the microphones, accelerometers, and touch screen touches;
\item Actuator demands generated by the robot's control system along with sampled position of joints, from which we can deduce the robot’s pose;
\item Synthesised speech and touch screen display outputs;
\item Periodically sampled battery levels;
\item Periodically sampled status of the robot's WiFi and Internet connectivity, and
\item The robot's position (i.e. x,y coordinates) within its working environment (noting that this data might be available from a tracking system in a `smart' environment, or if not, deduced via odometry from the robot's known start position at, say, its re-charging station and subsequent movements).
\end{enumerate}

The EBB should also record high level decisions generated by the robot’s AI (see the data flow in Fig. 1), and -- given that social robots are likely to accept speech commands -- we would, ideally, be able to record both the raw audio recorded by the microphones and the text sequence produced by the robot's speech recogniser.

\subsection{Process}

Conventionally accident investigation is performed in four stages: (1) information gathering, (2) fact analysis, (3) conclusion drawing and -- if necessary -- (4) implementation of remedies. Stage (2) fact analysis might typically take the form of causal analysis, and we propose to adopt here the method of why-because analysis developed by Ladkin et al. \cite{Lad01,lad05,sand12}.

\begin{figure}[ht]
\begin{center}

\includegraphics[scale=.4]{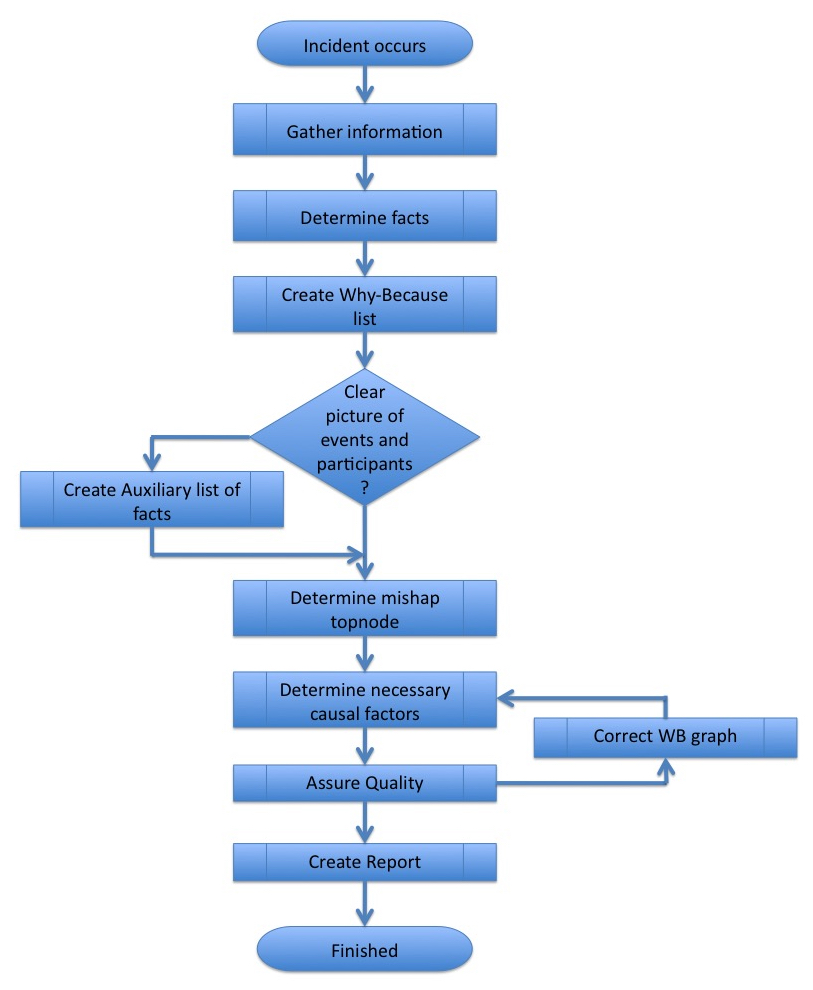}
\caption{An overview of Why-Because Analysis (adapted from Ladkin, 2005)} 
\label{fig:2}       
    
\end{center}
\end{figure}

Why-Because Analysis (WBA) is a method of causal analysis of complex socio-technical systems, and hence well suited to social robot accidents. WBA lends itself to a simple quality check: whether one event is a causal factor in another can be determined by applying a counter-factual test. The primary output of WBA is a Why-Because Graph (WBG), and a further advantage of WBA is that -- if necessary -- the correctness of the WBG can be formally proven. Fig. \ref{fig:2} shows, in flow chart form, the process of why-because analysis.

Let us elaborate briefly on some of the steps in Fig. \ref{fig:2}. `Gather information' requires collecting all available information on the accident. This will include witness testimony, records from the EBB, any forensic evidence, and contextual information on the setting of the accident. The next stage `determine facts' requires the investigation team to sift the information and glean the essential facts. Noting that witness testimony can be unreliable, any causal inferences from that testimony should ideally be corroborated with forensic evidence, so that -- even if not absolutely certain -- the team can have high confidence in those inferences. The third stage: `create why-because list' links the facts of events -- including both things that happened and things that should have happened but did not (unevents) -- to outcomes. If they give the team a clear picture of the sequence of events and participants in the accident then the team agree on the `mishap topnode(s)' of the why-before graph, i.e. the accident -- or perhaps multiple accidents.  Then the why-because graph is created, top-down. This is likely also to required several iterations and -- if necessary -- quality checking using counter-factual tests or formal proof of completeness. For a complete explanation of the method refer to the the introduction to WBA in \cite{sand12}.

\subsection{The application of the framework} 

To understand how this framework would operate in practice, we return to the fictional scenario outlined at the start of the paper. As described in the scenario, an elderly lady ‘Rose’ has a fall in her home. She is found, still on the floor, some time later by her neighbour, Barbara. Barbara calls for medical help as well as alerting Rose's family. Whilst Rose receives hospital treatment, an investigation team is formed, who begin to collect evidence for the investigation. Photos of Rose's flat are taken and information about her home set up is collected; for instance, Rose lives in a smart home with various sensors to detect her movements and communicate with the robot as necessary. Preliminary observation of the robot also reveals details about its design and use in the home. The robot can fetch drinks, provide reminders (such as for Rose to take medication) and act as an interface so that Rose can control her television and other devices through it with voice commands. The robot will also respond to commands sent via a smartphone/tablet application. 

Barbara, the paramedics and Rose herself are interviewed to provide witness statements. These statements combine with the initial observations to provide important early findings that shape the ongoing investigation. Of key importance are the following:  i) Rose didn't put on her fall alert bracelet on the day of the accident, and ii) From her condition (as observed by the paramedics) it seems that after her fall Rose was too weak to call out to the robot if it was more than two metres away from her. 

In her witness testimony Rose states that she had climbed on a chair and was reaching for something in an upper cupboard in her kitchen but then slipped and fell on the floor. She has no memory of what happened after that and does not recall where the robot was when she fell. Barbara states that she doesn't recall seeing the robot when she entered the flat, and feels that the shock of finding Rose on the floor meant she didn't notice much else around her. The paramedic states that he noticed the robot moving about around the flat -- between the living area and the kitchen. It didn't seem to be moving for the accomplishment of any particular action so he described the robot as acting ‘aimlessly’. 

The investigation team gather further information. They interview the manager of the retirement complex that Rose lives in; she provides details of the organisational structure of the complex including the infrastructure that enables residents to set up their homes as smart homes and have assistance robots. They also talk to others who saw Rose on the day of her accident. The last person to see Rose before her fall was Michelle, a cleaner who works for the retirement complex. Michelle saw Rose, whilst she was in Rose's flat for an hour to do her regular weekly clean. Michelle reported that Rose seemed very cheerful and chatty, and did not complain of feeling ill or mention any worries or concerns. Michelle said that she did her usual clean in its usual order as requested by the retirement complex:  collect up rubbish to take outside; wipe bathroom surfaces and floor; wipe kitchen work surfaces and clean floor; polish wooden surfaces; hoover carpeted areas; disinfectant wipes on door handles and all over the robot for infection control. When asked by the investigation team she said she thought the robot was in the living room for most of the time she was there but she couldn't really remember. She didn't notice anything unusual about what the robot was doing.
 
The team also gets a report from Rose's General Practitioner about her overall health status prior to the accident. This states that Rose had some limited mobility and needed to take daily medication to help her arthritis. She had also recently complained of forgetting things more and more easily. However she was generally healthy for her age and had no acute concerns.  

Finally, the team extracts data from the Ethical Black Box. These are in CSV format, and contain timestamped information regarding (i) the location and status of the robot and Rose/others within the apartment, (ii) actions undertaken by the robot and (iii) sampled records of all other robot inputs/outputs over the previous 24 hour period. It enables the team to conclude that the robot lost connection to the central `home connection hub' intermittently over the past 24 hours, coinciding with Rose's fall. In addition, processing of the camera feed and other sensors used for navigation appear to be producing erroneous data. The records showed no log of Rose's fall, but did log that the robot made a number of `requests for help' -- by speaking out loud -- regarding its inability to connect to the home connection hub.

Having collected and analysed all of the material, the investigation team identify key relevant factors. At the individual level, certain actions by Rose -- forgetting to put on her accident bracelet and reaching to a high cupboard -- certainly increased the safety risk. Aspects of the local environment are likely to have also contributed to this risk and influenced the technical problems that occurred -- for instance the repeated disinfecting of the robot, as required by the retirement complex, has almost certainly impaired its sensors. The robot's response to losing connection to the home hub, i.e. ‘asking for help’ was clearly not effective in getting the problem addressed, most likely because Rose did not understand the problem. 

\begin{figure}[ht]
\begin{center}
\includegraphics[scale=.3]{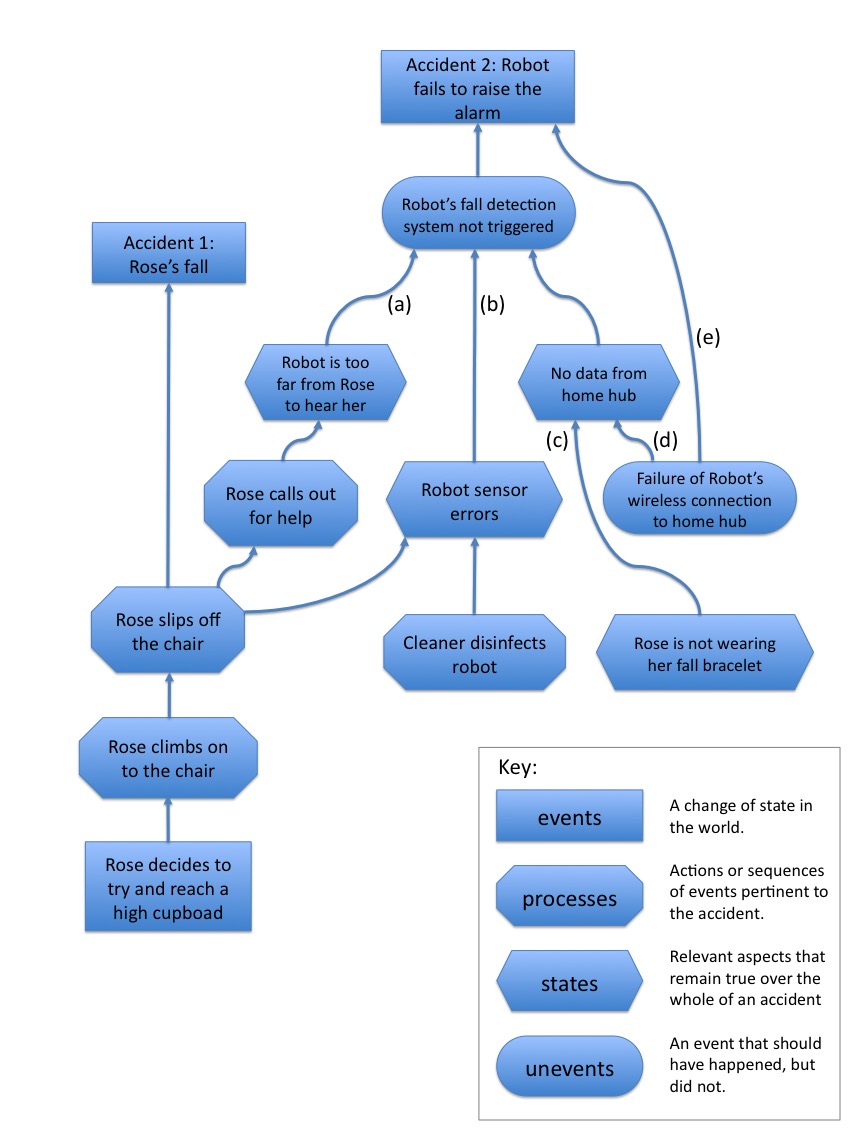}
\caption{Why-because graph for Rose's accident} 
\label{fig:3}       
\end{center}
\end{figure}

Concerning the robot’s standard functionalities, it failed to detect Rose's fall and therefore raise an alert following the fall. The robot's fall detection system relies, in part, on data collected by distributed sensors placed around the smart home. This data is delivered to the robot via the home connection hub, so the intermittent connectivity issues prevented the robot's fall detection functionality from operating as intended. The team make use of these key facts to construct the why-because graph shown in Fig. \ref{fig:3}.  

The first thing to note in Fig. \ref{fig:3} is that there are two mishap topnodes and hence two sub-graphs. On the left is the sub-graph showing the direct causal chain that resulted in Rose's fall (Accident 1), following her unwise decision to try and reach a high cupboard. The sub-graph on the right shows the chain of factors -- processes, states and unevents (things that should have happened but didn't) -- that together led to Accident 2: the robot failing to raise the alarm. 

The sub-graph leading to Accident 2 shows that the four ways in which the robot might have raised the alarm all failed, for different reasons. The first (a) is that the robot was too far away from Rose to hear her calls for help (most likely because the failure of its connection with the home hub means that the robot didn't know where she was). The second (b) is that the robot's sensors that should have been able to detect her fall (together with data from the smart environment) were damaged, almost certainly by cleaning with disinfectant, and the third (d) was the failure of the wireless communication between the robot and home hub, which meant there was no data from the home's smart sensors. A fourth reason (c) is due to two factors (i) Rose had forgotten to put on her fall alarm bracelet, but (ii) even if she had been wearing it the bracelet would have been ineffective as it too communicates with the robot via the home hub. The failure of communication between the robot and home hub is particularly serious because, as the graph shows, even if the first two pathways (a) and (b) to the robot's fall detection system had operated correctly the robot would still not have been able to raise the alarm, indicated by path (e). To use the Swiss cheese metaphor from section 2, over reliance on communication with the home hub represents a set of holes in the layers of safety which all line up.

The key conclusions from this analysis are that (i) the robot did \textbf{not} cause Rose's accident, (ii) the robot failed to raise the alarm following Rose's fall -- one of its primary safeguarding functions, and (iii) failures and shortcomings of the smart home's infrastructure contributed significantly to the robot's failure. The robot's failure might have had very severe consequences had a neighbour not called upon Rose and raised the alarm.

As a consequence of their investigation the team are able to to make a set of recommendations to prevent similar accidents happening in future. These recommendations are, in order of priority:
\begin{enumerate}
\item Equip the robot with a backup communications system, in case the WiFi fails. A recommended approach might, for instance, be to integrate a module allowing the robot to send text or data messages via a 3G GSM connection to the public cellular network.
\item Equally important is that if the robot detects a WiFi connectivity failure it should not be alerting it’s user (Rose) but instead sending an alert to a maintenance engineer via its backup communication system. 
\item Equip the home hub with the ability to send an emergency call directly -- via a landline for instance -- when the fall bracelet is triggered, so that this particular alarm is not routed via the robot.
\item Improve the sensitivity of the robot's microphones to increase their range.
\item Add a new function to the robot so that it reminds Rose to put on her fall bracelet every day.
\item Advise the cleaner not to use disinfectants on the robot.
\end{enumerate}

\section{Concluding discussion}

\subsection*{RoboTIPS}

The work of this chapter is part of five-year programme RoboTIPS: Responsible Robots for the Digital Economy\footnote{\url{https://www.robotips.co.uk/}}. RoboTIPS has several themes, two of which are of relevance to this paper. The first is the technical specification and prototyping of the EBB, including the development of model EBBs which will be used as the basis of implementations and trials by project industrial partners. There will be two model EBBs, one a hardware implementation and the other a software module, and their specifications, code and designs will be published as open source in order to support and encourage others to build EBBs into their robots. Ultimately we would like to see industry standards emerge for EBBs, noting that we would need separate specifications for different domains: one EBB standard for AVs, another for healthcare robots, a third for robot toys/educational robots, and so on.

Second we are designing and running three staged (mock) accidents, and we anticipate one in the application domain of assisted living robots, one in educational (toy) robots, and another for autonomous vehicles. We believe this to be the world's first experimental investigation of accident investigation in the field of social robots. For each of these staged scenarios we will be using real robots and will invite human volunteers to act in three roles, as 
\begin{enumerate}
\item subject(s) of the accident, 
\item witnesses to the accident, and as 
\item members of the accident investigation team.
\end{enumerate}
One key aim of these staged accidents is to trial, develop and refine the framework for social robot accident investigation outlined in this paper.  

Thus we aim to develop and demonstrate both technologies and processes (and ultimately policy recommendations) for effective social robot accident investigation. And as the whole project is conducted within the framework of Responsible Research and Innovation it is a case study in Responsible Robotics.

\subsection*{The Bigger Picture}

There are two important details that we omitted from the accident scenario outlined in section 1, then developed in section 5.3. The first is who needs to be on a robot accident investigation team. And the second -- and perhaps more fundamental question -- who do you call upon to investigate a social robot accident?

Concerning the makeup of a social robot accident investigation team, if we follow best practice it would be a multi-disciplinary team. One report, for instance, described multi-disciplinary teams formed to investigate sleep related fatal vehicle accidents as ``consisting of a police officer, a road engineer, a traffic engineer, a physician, and in certain cases a psychologist'' \cite{rad04}. Such teams did not require the involvement of vehicle manufacturers, but more recent fatal accidents involving AVs have needed to engage the manufacturer, to provide both expertise on the AV's autopilot technology and access to essential data from the vehicle's proprietary data logger \cite{ntsb16}. Robot accident investigations will similarly need to call upon the assistance of robot manufacturers, both to provide data logs and advice on the robot's operation. We would therefore expect social robot accident investigation teams to consist of (i) an independent lead investigator with experience of accident investigation, (ii) an independent expert on human-robot interaction, (iii) an independent expert on robot hardware and software, (iv) a senior manager from the environment in which the accident took place, and (v) one of the robot manufacturer's senior engineers. Depending on the context of the accident the team might additionally need, for instance, a (child-)psychologist or senior health-care specialist.

Consider now the question: who do you call when there has been a robot accident?\footnote{After the paramedics, that is.} At present there is no social robot equivalent of the UK Air Accident Investigations Branch\footnote{\url{https://www.gov.uk/government/organisations/air-accidents-investigation-branch}}, or Healthcare Safety Investigation Branch (HSIB)\footnote{\url{https://www.hsib.org.uk/}}. A serious AV accident would of course be attended by a police road traffic accident unit, although they would almost certainly encounter difficulties getting to the bottom of failures of the vehicle's autopilot AI. The US National Transport Safety Board\footnote{\url{https://www.ntsb.gov/Pages/default.aspx}} (NTSB) is the only investigation branch known to have experience of AV accidents, having investigated five to date (it is notable that the NTSB is the same agency responsible for air accident investigation in the US, and thus able to bring that considerable experience to bear on AV accidents).

For assisted living robots deployed in a care home settings, such as in our example scenario, accidents could be reported to both the Care Quality Commission\footnote{\url{https://www.cqc.org.uk/}} (CQC) -- the regulator of health and social care in England -- and/or the Health and Safety Executive\footnote{\url{https://www.hse.gov.uk/}} (HSE), since care homes are also workplaces. Again it is doubtful that either the CQC or HSE would have the expertise needed to investigate accidents involving robots. Accidents involving workplace assistant robots, or robots used in schools -- including near misses --  would certainly need to be reported to the HSE. It is clear that as the use of social robots in society increases, regulators such as the CQC and HSE will need to create robot accident investigation branches, as would the HSIB for surgical or healthcare robots. Even more urgent is the need to record all such accidents -- again including near misses -- so that we have, at the least, statistics on the number and type of such accidents.

Until such mechanisms exist, or for robot accidents in settings that fall outside those outlined here, the only recourse we would have is to contact the robot's manufacturer, thus underlining the importance of clear labelling of the robot’s make and model alongside contact details for the manufacturer of the robot itself\footnote{See EPSRC Principle of Robotics \#5 in \cite{boden17}.}. Even if the robot and its manufacturer does not yet have data logging technologies (such as the EBB) or processes for accident investigation in place, we would hope that they would take accidents seriously. A responsible manufacturer would both investigate the accident -- drawing in outside expertise where needed -- and effect remedies to correct faults. Ideally social robot manufacturers would adopt the data sharing philosophy that has proven so effective in aviation safety, summed up by the motto ``anybody’s accident is everybody’s accident''.

\begin{acknowledgement}
The work of this chapter has been conducted within EPSRC project Robo-TIPS, grant reference EP/S005099/1 RoboTIPS: Developing Responsible Robots for the Digital Economy.
\end{acknowledgement}

\bibliographystyle{plain}
\bibliography{RAIrefs}

\end{document}